\documentclass[runningheads]{llncs}
\usepackage{graphicx}
\usepackage{graphicx}
\graphicspath{ {pictures/} } 
\usepackage{color}
\usepackage{amsmath}
\usepackage{multicol}
\usepackage{footmisc}
\usepackage{epstopdf}
\usepackage{subcaption}
\def\acknowledgement{\par\addvspace{17pt}\small\rmfamily
\trivlist\if!\ackname!\item[]\else
\item[\hskip\labelsep
{\bfseries\ackname}]\fi}

\newenvironment{acknowledgements}{\begin{acknowledgement}}
{\end{acknowledgement}}

\begin{document}
\title{Fast Simulation of Crowd Collision Avoidance }
%
%
\author{John Charlton\inst{1}\orcidID{0000-0001-8402-6723} \and
Luis Rene Montana Gonzalez\inst{1} \orcidID{0000-0002-0511-0466} \and
Steve Maddock\inst{1}\orcidID{0000-0003-3179-0263} \and
Paul Richmond\inst{1}\orcidID{0000-0002-4657-5518}}
\authorrunning{J. Charlton et al.}
%
\institute{University of Sheffield, Sheffield S10 2TN, UK \email{\{j.a.charlton,lrmontanagonzalez1,s.maddock,p.richmond\}@sheffield.ac.uk} }
\maketitle              
\begin{abstract}
Real-time large-scale crowd simulations with realistic behavior, are important for many application areas. On CPUs, the ORCA pedestrian steering model is often used for agent-based pedestrian simulations. This paper introduces a technique for running the ORCA pedestrian steering model on the GPU. Performance improvements of up to 30 times greater than a multi-core CPU model are demonstrated. This improvement is achieved through a specialized linear program solver on the GPU and spatial partitioning of information sharing.  This allows over 100,000 people to be simulated in real time (60 frames per second). 

\keywords{Pedestrian Simulation  \and Real-time rendering \and GPU-computing}
\end{abstract}
%

\section{Introduction}

Crowd simulations are important for many applications, such as safety studies for communal transport hubs and flows within sports stadiums and large buildings \cite{xu_Crowd_2014}. Such simulations require believable dynamics that match observed behavior, including correct collision avoidance, or steering behavior. The Optimal Reciprocal Collision Avoidance (ORCA) algorithm \cite{berg_reciprocal_2011} is an agent-based solution that can simulate many real crowd behaviors. Currently, implementations of the ORCA algorithm have been made for single- and multi-core CPU. This paper presents a GPU implementation, supporting real-time simulations and interactivity for very large populations of order $5\times10^5$.

Computer models that contain inherent parallelism are suitable candidates for GPUs. This applies to agent-based pedestrian simulation models, where all agents follow the same rules. Using steering techniques that lend themselves well to implementation on GPU architecture can result in much faster performance \cite{barut_combining_2018,bleiweiss_multi_2009}. By increasing performance, greater numbers of people can be simulated and/or a more accurate, possibly more time-consuming, algorithm can be used for the simulation.

This paper presents a GPU implementation of the ORCA model for agent-based pedestrian simulation. We parallelize as much of the data and computation as possible, choosing data parallel algorithms and spatial partitioning to allow communication between people to provide speedup. Our solution makes use of a novel low-dimension linear program solver developed for the architecture of a GPU \cite{charlton_two-dimensional_2019}, and a grid-based spatial partitioning scheme of information transfer between GPU threads \cite{richmond_flame_2011}. Grid partitioned data structures are an efficient form of spatial partitioning on the GPU \cite{li_comparative_2013}. Our GPU implementation shows performance increases of up to 30 times over the original CPU multi-core version \cite{berg_reciprocal_2011,snape_optimal_2019} with these changes. In addition, it consistently outperforms the CPU version for sufficiently large amounts of people. 

The organization of the paper is as follows. Section \ref{sec:background} covers background information and related work. Section \ref{sec:algorithmology} explains in detail the implementation of the ORCA model on the GPU. Section \ref{sec:results} presents results and discussion of the multi-core CPU and GPU ORCA models. Finally, section \ref{sec:conclusion} gives the conclusions.

\section{Background} \label{sec:background}

Many types of models have been proposed to generate local pedestrian motion and collision avoidance \cite{pettre_motion_2008,bousseau_introduction_2017,thalmann_populating_2006}. The simplest separation of steering models is between continuum models and microscopic models. Continuum models attempt to treat the whole crowd in a similar way to a fluid, allowing for fast simulation of larger numbers of people, but are lacking in accuracy at the individual person scale \cite{narain_aggregate_2009}.  Moving part of the calculation to the GPU has shown performance improvements \cite{fickett_GPU_2007}. Overall, however, the model is not ideal for solving on the GPU due to the large sparse data structures. In comparison, microscopic models tend to be paired with a global path planner to give people goal locations and trajectories. Such models specify rules at the individual person scale, with crowd-scale dynamics being an emergent effect of the rules and interactions, and easily allow for non-homogeneous agents and behavior.

Popular microscopic models are cellular automata (CA), social forces\cite{helbing_social_1995} and velocity obstacles (VO) \cite{fiorini_motion_1998}. CA are popular due to the ability to reproduce observable phenomena \cite{blue_emergent_1998,blue_cellular_1999}, but a downside is the inability to reproduce other behaviors due to using discrete space. CA models are computationally lightweight and lend themselves well to specify certain complex behavior. However, CA pedestrian models tend to use discrete spatial rules, where the order of agent movements are sequential, which does not lend itself to parallelism and GPU implementations \cite{schonfisch_synchronous_1999}. Social forces models use a computationally lightweight set of rules that allows for crowd-scale observables such as lane formation. They are well suited to parallelizing on the GPU since all agents can be updated simultaneously, with good performance for many simulated people \cite{karmakharm_agent-based_2010,richmond_high_2008}. However, generated simulations can result in unrealistic looking motion and produce undesirable behavior at large densities.

Velocity obstacles (VO) work by examining the velocity and position of nearby moving objects to compute a collision-free trajectory. Velocity-space is analyzed to determine what velocities can be taken which do not cause collisions. 
VO models lend themselves to parallelization since agents are updated simultaneously and navigate independently of one another with minimal explicit communication. It tends to be more computational and memory intensive than social forces models, but the large throughput capability of the GPU for such parallel tasks make it a very suitable technique for GPU implementation.
Early models assumed that each person would take full responsibility for avoiding other people. Several variations include the reactive behavior of other models \cite{abe_collision_2001,kluge_recursive_2006,fulgenzi_dynamic_2007}. One example is reciprocal velocity obstacles (RVO), where the assumption is that all other people will take half the responsibility for avoiding collisions \cite{berg_reciprocal_2008,guy_clearpath_2009}. This model has been implemented on the GPU \cite{bleiweiss_multi_2009} and has shown credible speedup over the multi-core CPU implementation through use of hashing instead of naive nearest neighbor search. Group behavior has also been included in VO models \cite{he_dynamic_2016,yang_proxemic_2016} allowing people to be joined into groups. Such people attempt to remain close to other members of the group and aim for the same goal location.
A further extension is optimal reciprocal collision avoidance (ORCA). It provides sufficient conditions for collision-free motion. It works by solving low-dimension linear programs. Freely available code libraries have been implemented for both single- and multi-core CPU \cite{snape_optimal_2019}.

VO techniques are very suitable candidates for GPU implementation. The RVO model and implementation by Bleiweiss \cite{bleiweiss_multi_2009} show notable performance gains against multi-core CPU equivalent models. However, these methods must perform expensive calculations to find a suitable velocity. They tend to perform slower and are not guaranteed to find the best velocity. ORCA is deemed more suitable because of its performance relative to other VO models and collision-free motion, theoretically providing ``better" motion (i.e. less collisions). 

Linear programming is a way of maximizing an objective function subject to a set of constraints. For ORCA, linear programming is used to find the closest velocity to a person's desired velocity which does not result in collisions. It is important to choose a solver that is efficient on the GPU at low dimensions. A popular solver type is the Simplex method. This is best suited for large dimension problems and struggles at lower dimensions. The incremental solver \cite{seidel_small-dimensional_1991} is efficient at low dimensions but suffers on the GPU due to load balance: not all GPU threads have the same amount of computation, which reduces the performance on such parallel architecture.
The batch GPU two-dimension linear solver \cite{charlton_two-dimensional_2019} is an efficient way to solve the numerous linear problems simultaneously. We make use of this approach, demonstrating its use for large-scale pedestrian simulations.

\section{The Algorithm} \label{sec:algorithmology}
The proposed algorithm is based on the multi-core ORCA model and applies GPU optimizations. This section provides an overview of the algorithm as well as important changes and optimizations that need to be made to make the simulation efficient for running on the GPU. For more in-depth description of the ORCA algorithm, see the work of van den Berg et. al \cite{berg_reciprocal_2011}. The main changes are the use of an efficient linear program made for GPUs and an efficient method of communication between GPU cores for people to ``observe" properties of other people.

\begin{figure}
    \centering
    \begin{subfigure}{0.45\textwidth}
        \includegraphics[width=0.9\linewidth]{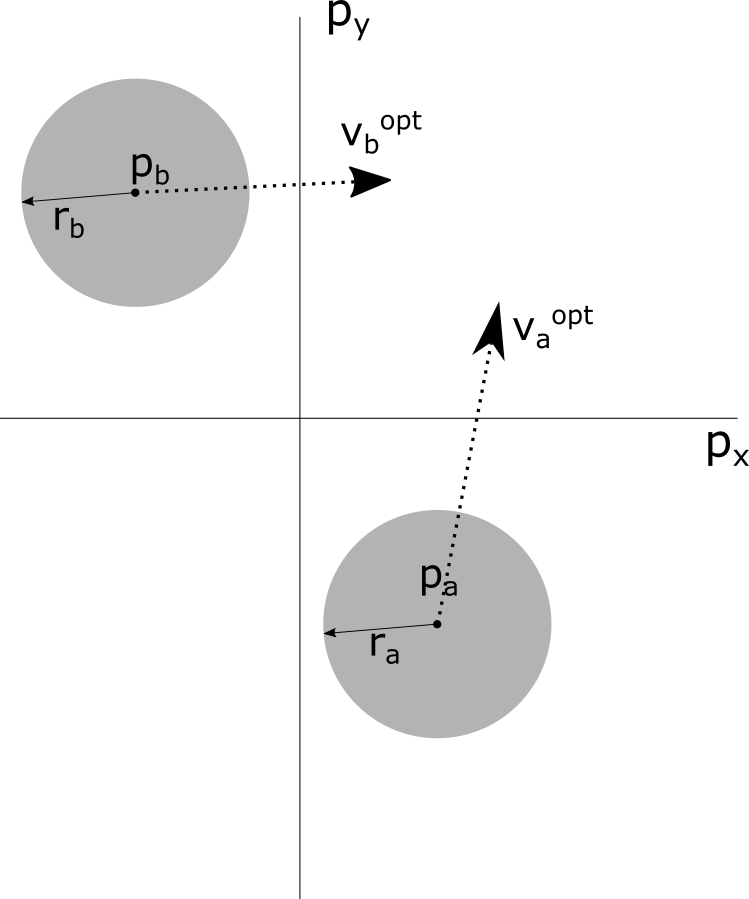}
        \caption{ }
    \end{subfigure}
     \begin{subfigure}{0.45\textwidth}
        \includegraphics[height=6cm]{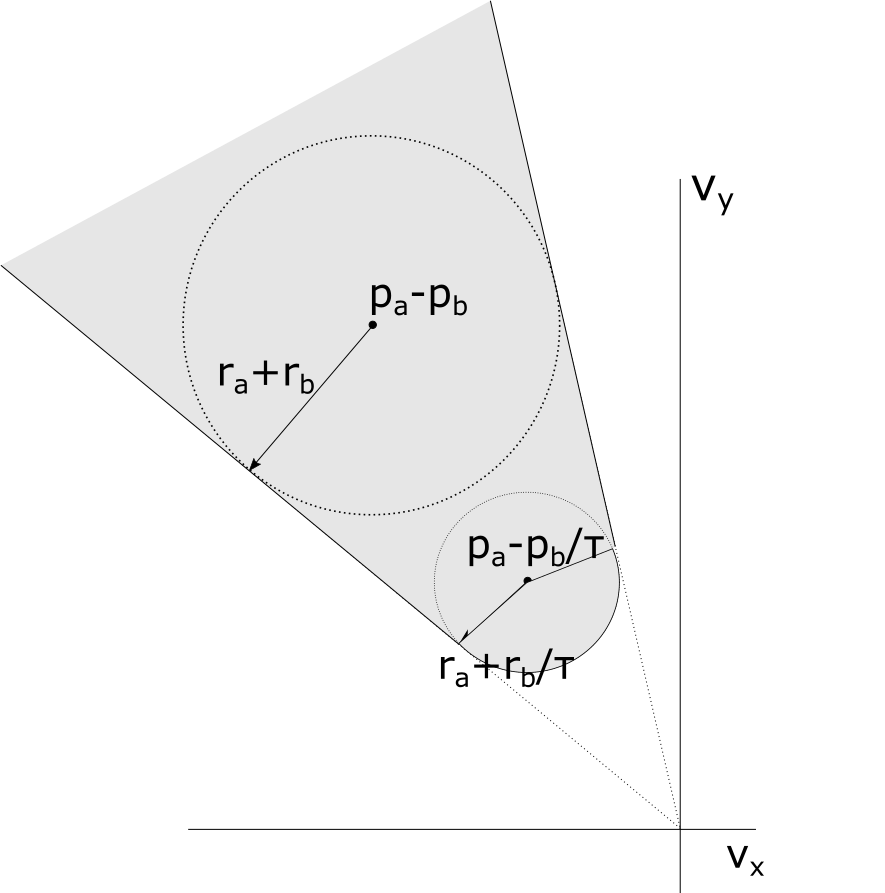}
        \caption{ }
    \end{subfigure}
     \begin{subfigure}{0.45\textwidth}
        \includegraphics[width=0.9\linewidth]{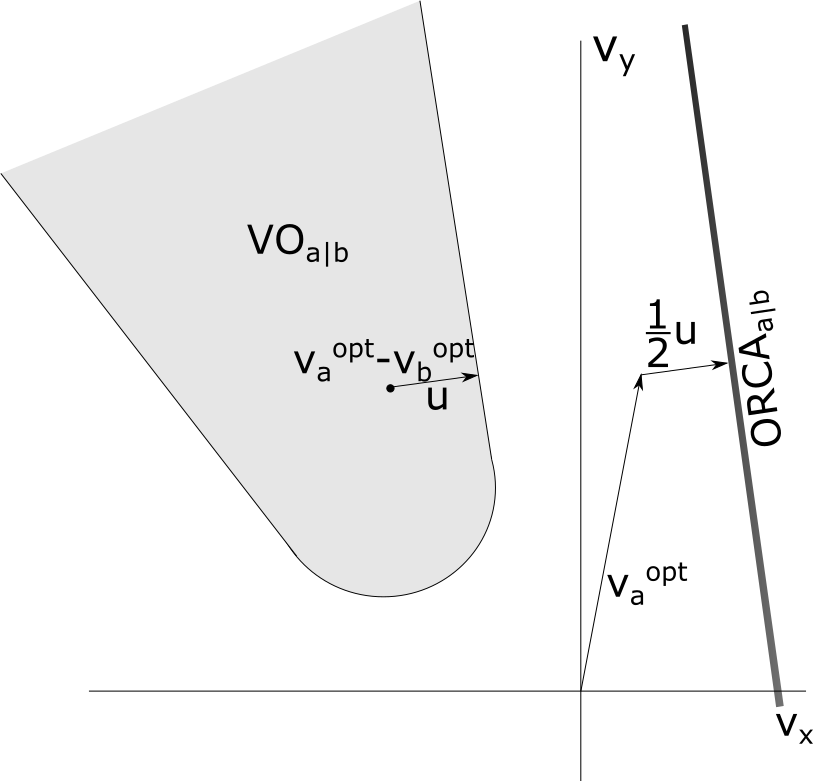}
        \caption{ }
    \end{subfigure}
    \begin{subfigure}{0.45\textwidth}
        \includegraphics[width=0.9\linewidth]{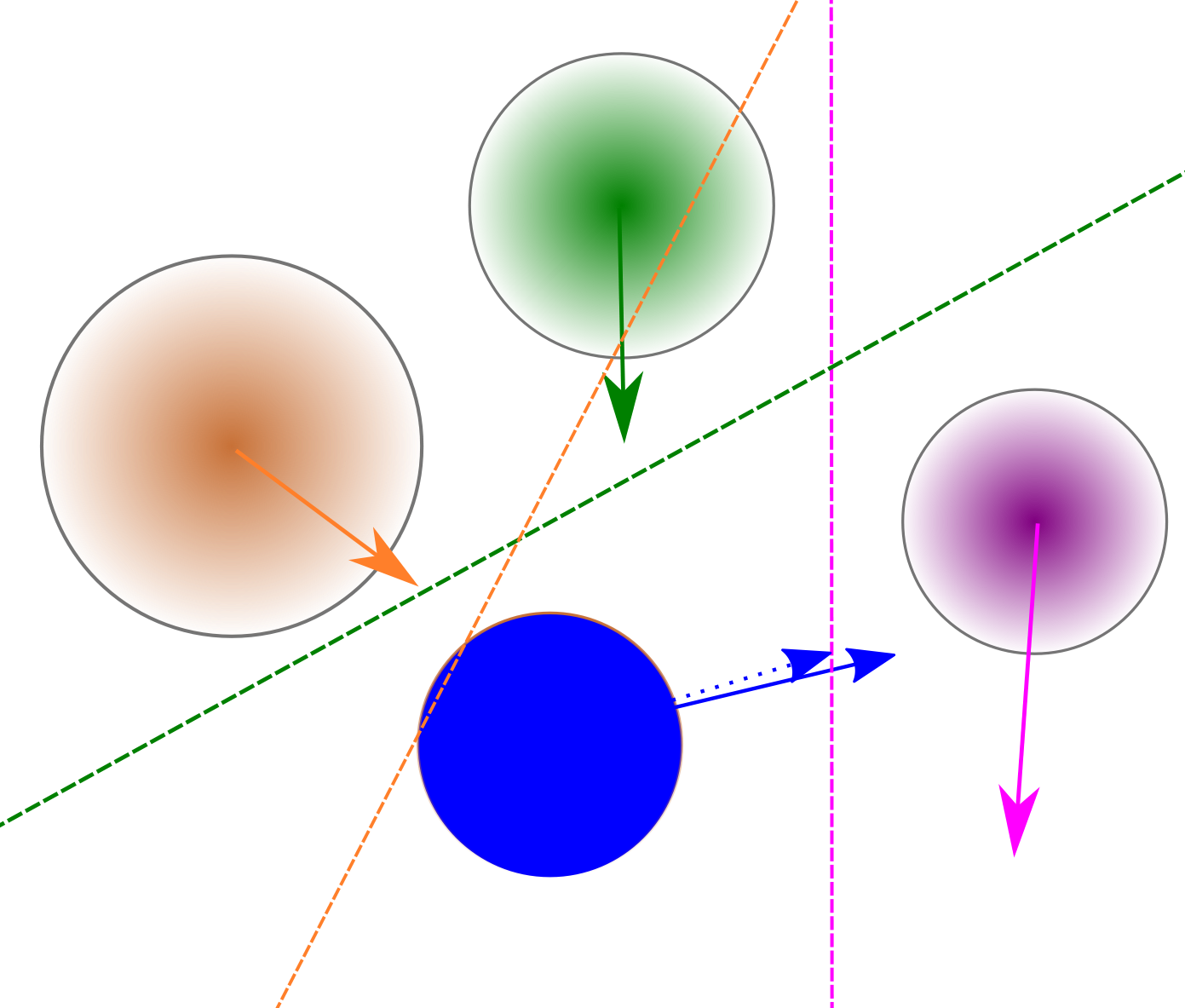}
        \caption{ }
        \label{fig:ORCAsolve}
    \end{subfigure}
    \caption{(a) A system of 2 people $a$ and $b$ with corresponding radius $r_a$ and $r_b$. (b) The associated velocity obstacle $VO_{a|b}$ in velocity space for a look-ahead period of time $\tau$ caused by the neighbor $b$ for $a$. (c) The vector of velocities $v_a^{opt} - v_a^{opt}$ lies within the velocity obstacle $VO_{a|b}$. The vector $u$ is the shortest vector to the edge of the obstacle from the vector of velocities. The corresponding half-plane $ORCA_{a|b}$ is in the direction of $u$, and intersects the point $v_a^{opt} + u$. (d) A view of a blue agent and its neighbors, as well as the generated half-planes caused by the neighbors interacting with the blue agent. The solid blue arrow shows the desired velocity of the blue agent. The dotted blue arrow is the resulting calculated velocity that does not collide with any neighbor in time $\tau$.}
    \label{fig:ORCAlines}
\end{figure}

As an overview to the ORCA model, each person in the model has a start location and an end location they want to reach as quickly as possible, subject to an average speed and capped maximum speed. For each simulation iteration, each agent ``observes" properties of nearby people, namely radius, the current position and velocity. For each nearby agent a half-plane of restricted velocities is calculated (figure \ref{fig:ORCAlines}). By selecting a velocity not restricted by this half-plane, the two agents are guaranteed to not collide within time $\tau$, where $\tau$ is the \textit{lookahead time}, the amount of forward time planning people make to avoid collisions. By considering all nearby agents, the set of half-planes creates a set of velocities that, if taken, do not collide with any nearby agents in time $\tau$. The agent then selects from the permissible velocities the one closest to its desired velocity and goal. Figure \ref{fig:ORCAsolve} shows the resulting half-planes caused by neighboring agents on an example setup, and the optimal velocity that most closely matches the person's desired velocity.


It is possible that the generated set of half-planes does not contain any possible velocities. Such situations are caused by large densities of people. The solution is to select a velocity that least penetrates the set of half-planes induced by the other agents. In this case, there is no guarantee of collision-free motion. 

The computation of velocity subject to the set of half-planes is done using linear programming. The problem for the linear program is defined with the constraints corresponding to the half-plane $ORCA_{a|b}$ of velocities, attempting to minimize the difference of the suitable velocity from the desired velocity. Since each agent needs to find a new velocity, there is a linear problem corresponding to each agent, each iteration. The algorithm used to solve this is the batch-GPU-LP algorithm \cite{charlton_two-dimensional_2019}. It is an algorithm designed for solving multiple low-dimensional linear programs on the GPU, based on the randomized incremental linear program solver of Seidel \cite{seidel_small-dimensional_1991}. 

This batch-LP solver works by initially assigning each thread to a problem (i.e. one pedestrian). Each thread must solve a set of half-plane constraints, subject to an optimization function. Respectively, these are that the person should not choose a velocity that collides with other people, and the person wants to travel as close to their desired velocity as possible.

Each half-plane constraint is considered incrementally. If the current velocity is not satisfied by the currently considered constraint a new valid velocity is calculated. The calculation of a new velocity is one of the most computationally expensive operations. It is also very branched, as only only some of the solvers require a new valid velocity and others can maintain their current value. This branching calculation causes the threads that do no need to perform a calculation to remain idle while the other threads perform the operation. This is an unbalanced workload on the GPU device and can vastly reduce the throughput as many threads do not perform any calculations, exacerbated by the fact that those threads performing the operation must take a lot of time to complete the operation.

The implementation of this calculation uses ideas from cooperative thread arrays \cite{wang_gunrock_2016} to subdivide the calculation into \textit{``work units"}, blocks of equal size computation. These work units can be transferred to and computed by different threads, allowing for a balanced work load and good performance. If the thread does not need to compute a new velocity, then it can aid in another problem's calculation. This algorithm shows performance improvements over state-of-the-art CPU LP solvers and other GPU LP solvers \cite{charlton_two-dimensional_2019}. 

\begin{figure}
    \centering
    \includegraphics[height=3.5cm]{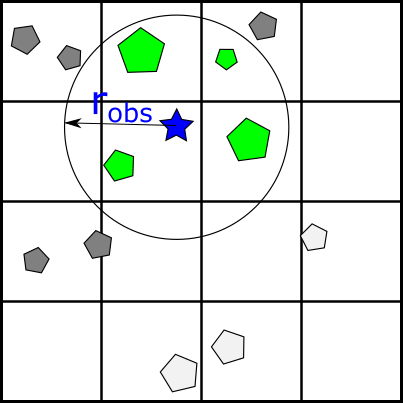}
    \caption{FLAME message partitioning. The simulation is discretized into spatial bins and people save their message to the corresponding bin. For a given person (blue star), it does not read messages of those in non-neighboring bins (white pentagons). For those within the same or neighboring partitioning bins, it calculates whether they are within the observation radius $r_{obs}$. If not, they are ignored (grey pentagons). If they are within the observation radius (green pentagon) the person is aware of them and will attempt to avoid them accordingly, by generating corresponding ORCA half-planes of valid velocities.}
    \label{fig:messages}
\end{figure}

The other main improvement is concerning the communication between people. Some information must be observed by people in the model. Examples are the radius, speed and position of others nearby. In order to communicate this information between people we use the idea of messages from the FLAME GPU framework \cite{richmond_flame_2011}, which is demonstrated in figure \ref{fig:messages}. Each agent creates a message which contains information on observables about themselves. Each message is assigned a spatial location equal to the position of the agent in the simulation. These messages are organized into spatial bins. Each agent will then read the messages from its associated bin, and those neighboring. This method is far faster than a brute-force read all approach for large spaces and many people/messages. The associated overhead in organizing messages into bins outweighs the cost of reading all messages and discarding those far away. A possible alternate implementation, that is used by the CPU model, uses a KD-tree spatial partitioning. It is expected, from the work of Li and Mukundan \cite{li_comparative_2013}, that this grid based spatial partitioning is faster than a KD-tree implementation on the GPU.

\section{Results} \label{sec:results}
This section presents the results of two experiments. The first experiment is composed of two test cases to demonstrate the appearance and correctness of the model. The first test case is a two-way crossing and the second test case is an eight-way crossing. The second experiment demonstrates the performance compared to the equivalent multi-core CPU version \cite{berg_reciprocal_2011,snape_optimal_2019}.

For the first experiment, all the test cases are set up in a similar way. Multiple associated start and end regions are chosen, such that people are spawned in a start region with a target in an associated end region. Random spawn locations are chosen so that there is no overlap with other people within a certain time period based on person size and speed. Within a simulation, each agent has a goal location to aim for. The agent's velocity is in the direction of the goal location, scaled to the walking speed. Once a person reaches the goal location they are removed from the simulation. Once all people have reached their goal the simulation is ended.

\begin{figure}
\centering
    \begin{subfigure}[t]{1\textwidth}
      \includegraphics[width=1\textwidth]{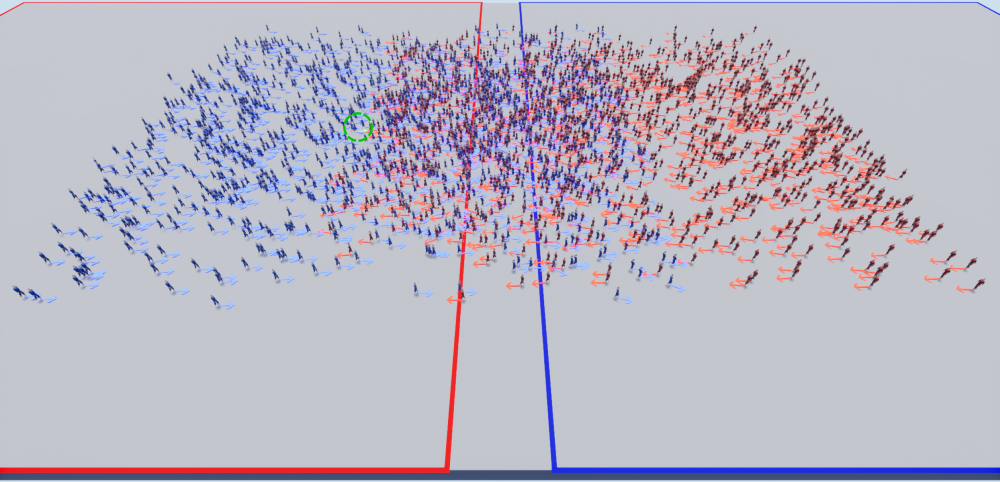}
    \end{subfigure}%
    \begin{subfigure}[t]{0.4\textwidth}
      \hspace{-1.02\textwidth}
      \includegraphics[width=1\textwidth]{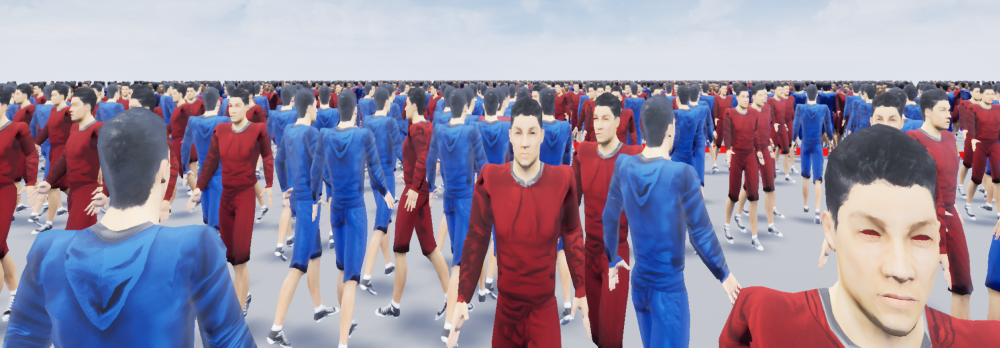}
    \end{subfigure}
    \caption{Visualization of 2,500 people in Unreal. Two crowds navigate past each other, one heading from left to right and the other heading from right to left. Left: scene view from above. Colored arrows show the direction of travel. One pedestrian is highlighted with a green circle; Inset: view from the perspective of the pedestrian in the green circle}
    \label{fig:vis1}
\end{figure}

The first test case was a 2-way crossing, with the two crowds attempting to pass amongst each other to reach their destination. Two variations of this were simulated. The first involves all people with the same size and speed parameters. The second version varies the size and speed parameters of each individual. Figure \ref{fig:vis1} shows the first variation for $2.5\times10^3$ people. The starting region of one group of people is the same area as the goal region of the other, forcing the two groups to navigate past each other. All agents have the same parameters, namely radius$=0.5\textrm{m}$, desired speed $=1.0\textrm{m/s}$ and maximum speed$=1.33\textrm{m/s}$. Various expected behaviors such as lane formation can be observed. The visualizations of the results are done by saving the agent data for each simulation step to a binary file and passing this to the Unreal engine.

\begin{figure}
\centering
\begin{subfigure}[t]{1\textwidth}
  \includegraphics[width=1\textwidth]{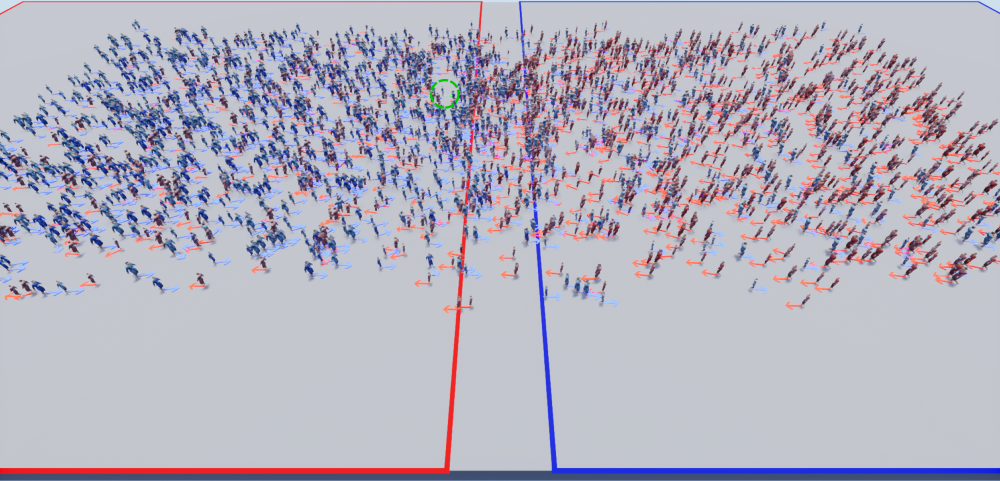}
\end{subfigure}%
\begin{subfigure}[t]{.4\textwidth}
  \hspace{-1.02\textwidth}
  \includegraphics[width=1\textwidth]{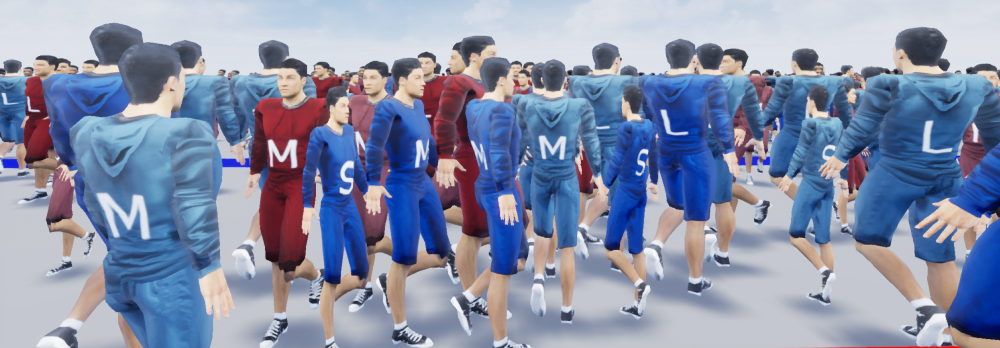}
\end{subfigure}
\caption{Visualization of 2,500 people in Unreal. Two crowds navigate past each other, one heading from left to right (blue clothes) and the other from right to left (red clothes). People are color coded according to their maximum speeds (using three shades of red or blue, respectively) and have varying radii (indicated by their actual size and also using S, M and L on their tops). Top: scene view from above; One pedestrian is highlighted with a green circle; Inset: view from the perspective of the pedestrian in the green circle}
\label{fig:vis2}
\end{figure}

In the second variation of the first test case, people have different sizes and different maximum speeds. Figure \ref{fig:vis2} illustrates this. In this example all people have an equal chance of being of radius $0.5\textrm{m}$, $0.75\textrm{m}$ or $1\textrm{m}$ (shown by person size in the figure, as well as S, M and L on their tops) and, independently, an equal chance of a desired speed of $1\textrm{m/s}$, $1.33\textrm{m/s}$ or $2\textrm{m/s}$. The maximum speed is adjusted to be $125\%$ of the desired speed. In figure \ref{fig:vis2}, people moving in the $x$ direction (left to right) have red tops, and people moving in the negative $x$ direction (right to left) have blue tops. Brighter shaded tops indicate the largest desired speed, $2\textrm{m/s}$, and the darkest tops indicate the slowest speed, $1\textrm{m/s}$.

\begin{figure}
\centering
  \includegraphics[width=.9\textwidth]{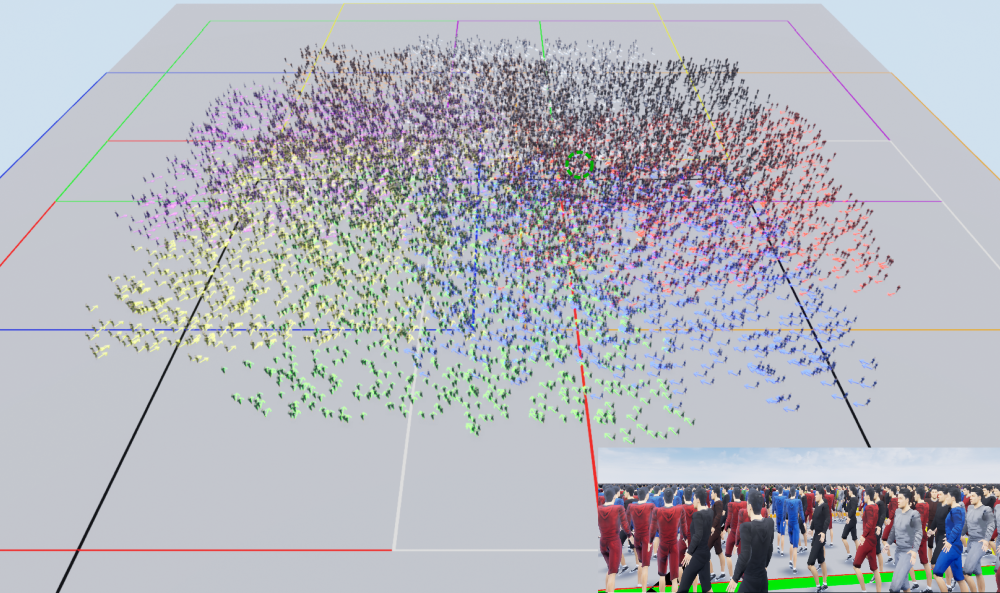}
    \caption{Visualization of 10,000 people in Unreal. Eight crowds attempting to navigate to the opposite end of the environment. Different colors are used for each crowd. Top: scene view from above; One pedestrian is highlighted with a green circle; Inset: view from the perspective of the pedestrian in the green circle}
\label{fig:vis3}
\end{figure}

The second test case was an 8-way crossing, visualized in figure \ref{fig:vis3}. Each crowd must navigate $135\deg$ across the environment, resulting in a vortex-like pattern around the center.

\begin{figure}
    \centering
    \includegraphics[width=.88\linewidth]{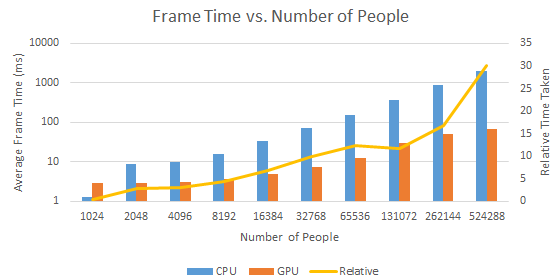}
    \caption{Frame time (in ms) for multi-core CPU and GPU ORCA models with varying numbers of people. Logarithmic scale on primary (left) vertical axis. Relative timing is given on the secondary (right) vertical axis, in linear scale. Simulation time only without visualisation of the pedestrians.}
    \label{fig:frameTime}
\end{figure}

The second experiment was designed to test the performance of the GPU implementation in comparison to the multi-core CPU implementation. Figure \ref{fig:frameTime} shows the results that varying numbers of people have on the frame time. Various test cases (e.g. 2-way and 8-way crossings) with different agent parameters were run, and the timings averaged between them. In this experiment no visualisation was used so as to ensure the timings were due to the algorithm only. The GPU solution gives speed increases of up to 30 times compared to the multi-core CPU implementation. Results for the single-core CPU version are not given as for any sizeable number of agents the multi-core CPU implementation always outperforms the single-core CPU implementation. This is due to better utilization of the CPU device. The colored bars of figure \ref{fig:frameTime} correspond to the primary (left) vertical axis, which uses a logarithmic scale. The relative time taken between the charts corresponds to the secondary (right) vertical axis, with linear scale.

The results show that the speed increases proportionally to the number of people. Greater relative speed-up occurs for even larger numbers of people, but the time taken per frame is below real time. The GPU simulations ran at close to 30 frames a second (33ms per frame) for up to $5\times10^5$ agents. The CPU version performs better for smaller number of agents, with a crossover occurring at approximately  $2\times10^3$ agents. This is due to the GPU device not being fully utilized for smaller simulations and the reduced throughput being outperformed by the CPU. The experiments were run on an NVIDIA GTX 970 GPU card with 4GB dedicated memory and a 4-core/8-thread Intel i7-4790K with 16 GB RAM. The GPU was connected by PCI-E 2.0. The GPU software was developed with NVIDIA CUDA 8.0 on Windows 10. On the GPU tested, there was a limit on the amount of usable memory of 4GB, which corresponded to approximately $5\times10^5$ people. It is expected that relative performance increases will continue to be obtained for larger numbers of people for the GPU implementation for GPUs with larger memory capacity.

\section{Conclusions} \label{sec:conclusion}
We have introduced a GPU-optimized version of the ORCA model. It shows substantial performance increases for large numbers of people compared to the multi-core CPU version. We demonstrated the performance gains through real-time visualizations that would not be possible on similar level CPU hardware.

Our model is currently limited in the number of people in the simulation size due to GPU memory. The models use large amounts of memory for storing the ORCA half-planes of each person. Memory usage could be reduced by considering fewer people. This would reduce the memory of each person but may result in less realistic motion with greater chance of collisions.  A solution to the lack of memory is with Maxwell and later architectures, which can use managed memory \cite{nvidia_tuning_2018} to page information from CPU to GPU on demand.  This would allow for many more people to be simulated, up to the computer's RAM capacity. It is expected that greater relative speedups between multi-core CPU and GPU will continue to be obtained for even larger amounts of simulated people.

It is expected that the more computationally expensive steering models would include more realistic motion such as side-stepping, more realistic densities, and less probability of collisions. In comparison, it is expected that the model in this paper would have greater performance and larger numbers of simulated people. 

The current work involves writing the data from the simulation to a file before visualization using Unreal. The data is copied from the GPU to the CPU, then loaded into Unreal and copied back to the GPU in Unreal for visualization. This is expensive. Future work will look at how to use the Unreal engine to visualize a simulation as it is calculated, which could be done by sharing GPU buffer information between the simulation program and the Unreal Engine. 

\begin{acknowledgements}
This research was supported by the Transport Systems Catapult and the National Council of Science and Technology in Mexico (Consejo Nacional de Ciencia y Tecnolog\'ia, CONACYT).
\end{acknowledgements}

%
%
%
\bibliographystyle{splncs04} 
\bibliography{MyLibraryBBT}

\end{document}